\documentclass[lettersize,journal]{IEEEtran}
\usepackage{amsmath,amsfonts}
\usepackage{algorithmic}
\usepackage{algorithm}
\usepackage{array}
\usepackage[caption=false,font=normalsize,labelfont=sf,textfont=sf]{subfig}
\usepackage{textcomp}
\usepackage{stfloats}
\usepackage{url}
\usepackage{verbatim}
\usepackage{graphicx}
\usepackage{cite}

\usepackage{booktabs}
\usepackage{multirow}
\usepackage{multicol}
\usepackage{xcolor}

\begin{document}

\title{Physics-Driven Local-Whole Elastic Deformation Modeling for Point Cloud Representation Learning}

\author{Zhongyu Chen, Rong Zhao, Xie Han, Xindong Guo, Song Wang, Zherui Qiao
\thanks{Corresponding author: Xie Han.}
\thanks{Zhongyu Chen, Rong Zhao, Xie Han, Xindong Guo, Song Wang, Zherui Qiao are with the School of Computer Science and Technology, North University of China,Shanxi Provincial Key Laboratory of Machine Vision and Virtual Reality, Taiyuan, 030051, China.}}

\markboth{Journal of \LaTeX\ Class Files,~Vol.~14, No.~8, August~2021}%
{Chen \MakeLowercase{\textit{et al.}}: Physics-Driven Local-Whole Elastic Deformation Modeling for Point Cloud Representation Learning}


\maketitle

\begin{abstract}
Existing point cloud representation learning methods primarily rely on data-driven strategies to extract geometric information from large amounts of scattered data. However, most methods focus solely on  the spatial distribution features of point clouds while overlooking the relationship between local information and the whole structure, which limits the accuracy of point cloud representation. Local information reflect the fine-grained variations of an object, while the whole structure is determined by the interaction and combination of these local features, collectively defining the object's shape. In real-world, objects undergo deformation under external forces, and this deformation gradually affects the whole structure through the propagation of forces from local regions, thereby altering the object's geometric features. Therefore, appropriately introducing a physics-driven mechanism to capture the topological relationships between local parts and the whole object can effectively mitigate for the limitations of data-driven point cloud methods in structural modeling, and enhance the generalization and interpretability of point cloud representations for downstream tasks such as understanding and recognition. Inspired by this, we incorporate a physics-driven mechanism into the data-driven method to learn fine-grained features in point clouds and model the structural relationship between local regions and the whole shape. Specifically, we design a dual-task encoder-decoder framework that combines the geometric modeling capability of data-driven implicit fields with physics-driven elastic deformation. Through the integration of physics-based loss functions, the framework is guided to predict localized deformation and explicitly capture the correspondence between local structural changes and whole shape variations. Experimental results show that our method outperforms existing approaches in object classification and segmentation, demonstrating its effectiveness.
\end{abstract}


\section{Introduction}
\IEEEPARstart{P}{oint} clouds can accurately describe the geometric shape and spatial distribution of objects. They are widely used in tasks such as 3D shape analysis \cite{yu2024pedestrian}, object detection \cite{song2024robustness}, and segmentation \cite{he2025segpoint,deng2024banana}. The success of these tasks depends on the accurate modeling of point cloud features, while the irregularity and sparsity of point clouds make learning their effective representations a challenging but essential task.

\begin{figure}[t]
	\centering
	\includegraphics[width=\linewidth]{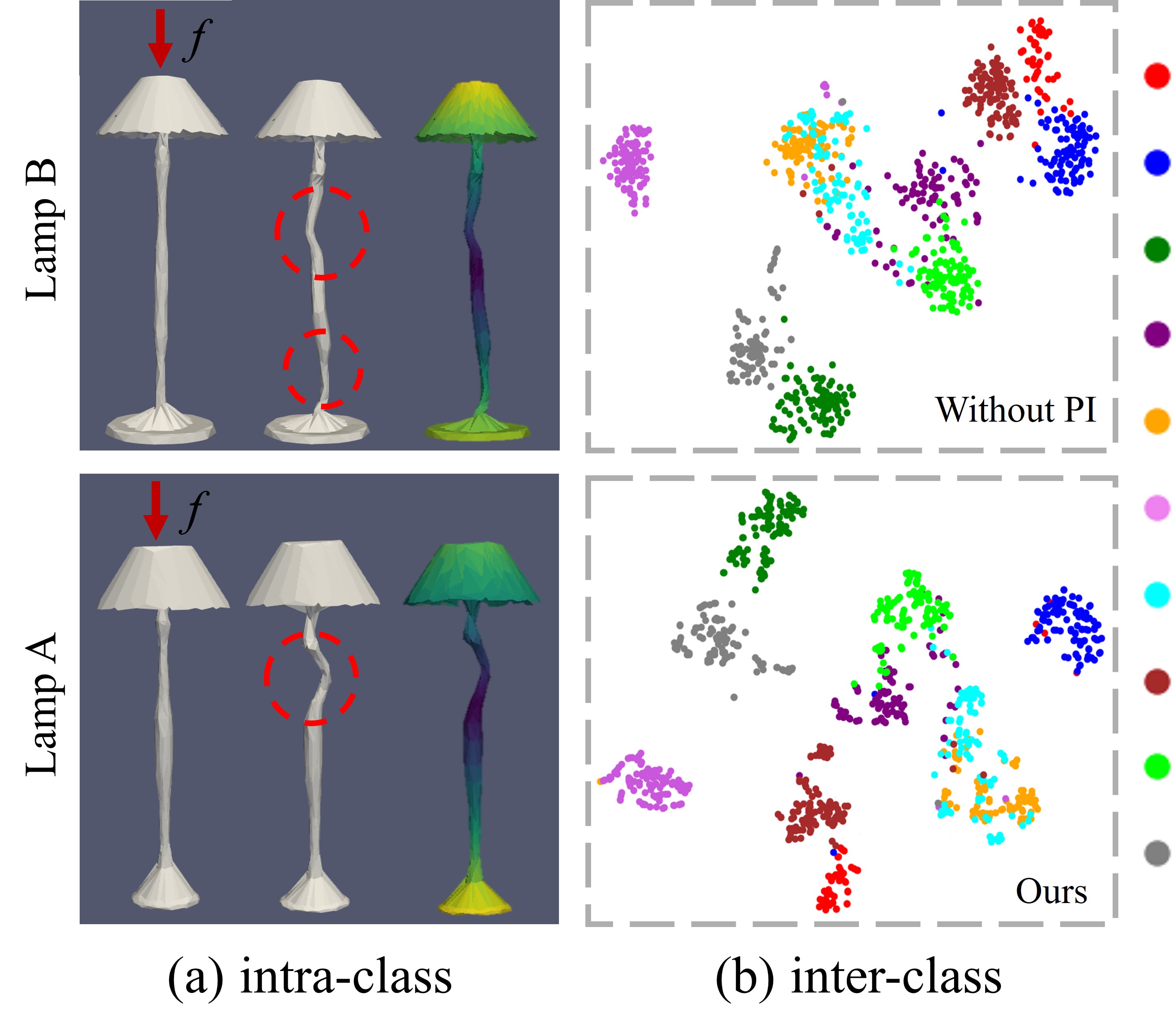}
	\caption{(a) Intra-class deformation and force distribution under an external force $f$ for Lamp A and B (colors indicate force magnitude; highlighted regions show significant deformation). (b) Inter-class clustering visualization in the embedding space across multiple categories, comparing models without and with Physics-Informed (PI) modeling. Note that (a) and (b) correspond to different tasks and the color encodings are not shared.}
	\label{fig1}
	\vspace{-1em}
\end{figure}

Existing methods primarily rely on data-driven approaches to capture geometric distributions of point clouds. In supervised point cloud representation learning, typical network architectures, such as PointNet \cite{qi2017pointnet}, use shared MLPs to process unordered point sets and directly learn the feature representations of each point in the point cloud to capture the whole geometric distribution. Building on this, PointNet++ \cite{qi2017pointnet++} aggregates local features hierarchically within fixed spatial neighborhoods via layered sampling and grouping, whereas DGCNN \cite{wang2019dynamic} dynamically constructs a $k$-NN graph in the feature space and learns with EdgeConv, allowing neighborhood relations to adapt as features are updated. In unsupervised point cloud representation learning, generative methods \cite{zhang2025point,zhang2022masked,lin2024patchmixing,han2023graph} learn latent representations of the data by reconstructing the geometric structure of the point cloud using generative models such as autoencoder (AE) \cite{liou2014autoencoder}, generative adversarial networks (GAN) \cite{creswell2018generative}, or autoregressive (AR) \cite{lin2024patchmixing}. Discriminative methods \cite{xie2020pointcontrast,huang2021spatio,wu2023self,li2024pointsmile} typically construct tasks by comparing feature differences between different augmented samples or categories, focusing more on learning the whole distribution patterns \cite{han2019multi,rao2020global,han2025cross} of point clouds in the feature space. Despite the significant progress made by these methods in point cloud representation learning, they often overlook the interaction between local information and whole structure, failing to explicitly model their interrelationship.

Specifically, existing methods often expand local information to the global level through multi-scale operations (e.g., the hierarchical structure of PointNet++) or max-pooling operations (e.g., global feature aggregation in PointNet). This process of aggregating information solely in the feature space does not explicitly model and capture the interrelationship between local and whole structures. For example, DGCNN captures local geometric structures using dynamic graph convolutional networks, but its layer-by-layer expansion of local information remains a passive, lacking explicit modeling of the relationship between part and whole structural features.

In nature, the shape of objects is often governed by the forces acting upon them. When an object is subjected to external forces, slight deformations first occur in its local regions. These deformations propagate through structural connections, causing global shape changes. This transition from ``local stress" to ``global response" reveals the inherent coupling relationships among the parts of an object. Meanwhile, recent research \cite{kadambi2023incorporating} has shown that physical modeling not only provides interpretability and inductive capability, but can also be embedded into deep networks as structural constraints to enhance the performance of data-driven methods on downstream detection tasks. As shown in Figure~\ref{fig1}, incorporating the physics-driven approach enables more effective discrimination of fine-grained features both intra-class and inter-class. Inspired by this, we incorporate a physics-driven mechanism into the data-driven method to extract local and whole structural shape relationships from object deformations. Specifically, we design a dual-task encoder-decoder framework. The encoder is responsible for learning the geometric information and physical properties of object from both the point cloud data and its corresponding tetrahedral mesh representation. The learned features are then input into two decoders: one decoder models the whole geometric shape of the point cloud using implicit fields, while the other decoder uses data-fidelity loss function and physics-informed loss function to predict local deformations, learning the force propagation relationship between local and whole structures. The main contributions of our method are as follows:

$\bullet$ We propose a novel self-supervised learning method that combines physical laws with geometric modeling and exploits force propagation to capture the relationships between local features and whole structures in point clouds.

$\bullet$ We designed two key loss functions: a data-fidelity loss that aligns the predicted displacement field for geometric consistency, and a physics-informed loss that enforces force equilibrium for physical consistency. Together, they regularize deformation behavior in point clouds.

$\bullet$ We demonstrate the effectiveness of our method through extensive evaluations on multiple downstream tasks, including object classification and segmentation. Additionally, the incorporation of physical information enhances the understanding of 3D point clouds.

\section{Related Work}
\subsection{Point Cloud Representation Learning}
In recent years, the advancement of deep learning techniques has enabled direct utilization of point cloud data for learning intrinsic feature representations of shapes. Early methods, such as PointNet \cite{qi2017pointnet}, used shared MLPs and max pooling operations to learn global features of point clouds. Subsequently, PointNet++ \cite{qi2017pointnet++} introduced a hierarchical structure and local feature learning mechanisms to address the issue of PointNet's insensitivity to local features in point clouds. DGCNN \cite{wang2019dynamic} further improves local feature extraction by dynamically constructing graphs based on point neighborhoods. Similarly, recent studies \cite{qian2022pointnext,ran2022surface,li2023pillarnext} have focused on learning local feature information from point clouds, aiming to extract as many semantic features as possible. However, the methods often treat point clouds as static geometric distributions, overlooking the explicit relationship between local and whole structures. This limitation, especially in the case of object deformation, makes it difficult to capture the impact of local changes on the whole structure. Against this backdrop, implicit field modeling \cite{yan2023implicit,WANG2025103120} has been gradually introduced into point cloud learning, representing the geometric structure of point clouds through continuous functions. This representation inherently provides spatial continuity and resolution adaptivity, making it well-suited for capturing the continuous relationship between local and whole structures. It also offers a stronger geometric foundation for subsequent unsupervised modeling.

\subsection{Unsupervised Point Cloud Representation Learning}
Nowadays, numerous unsupervised learning methods have emerged. Generative methods learn features through self-reconstruction \cite{wang2021unsupervised,zhang2025point,LI2023109200,yang2018foldingnet,han2023graph,tsai2022self}. For example, HC-VAE \cite{LI2023109200} introduces a hierarchical consistency mechanism to progressively generate high-quality point clouds. FoldingNet \cite{yang2018foldingnet} reconstructs arbitrary point clouds using a graph-based encoder and a folding-based decoder. Recently, recovering missing parts from masked inputs \cite{zhang2022masked,lin2024patchmixing,zhang2025point,zhang2023pointvst,10250984} has also become a prominent research focus. MaskSurf \cite{zhang2022masked} designs a transformer-based encoder-decoder network to estimate the position of the masking point and its corresponding normal vector. Point-DAE \cite{zhang2025point} reconstructs various corrupted point clouds by decomposing the reconstruction of a complete point cloud into detailed local patches and a coarse global shape, thereby alleviating the issue of positional leakage in the reconstruction process. 

Discriminative methods guide the model to learn meaningful feature representations by designing tasks that focus on the feature differences between samples. Some methods employ pretext tasks to learn point cloud representations \cite{sauder2019self,sharma2020self,xie2020pointcontrast,wu2024cross,malla2023clrgamcontrastivepointcloud}. Jigsaw3D \cite{sauder2019self} uses 3D puzzles as a self-supervised learning task, allowing the model to learn how to correctly assemble object parts. Poursaeed et al. \cite{poursaeed2020self} define the estimation of point cloud rotation angles as a pretext task. Recently, contrastive learning methods \cite{huang2021spatio,li2023tothepoint,xie2020pointcontrast,11153700} have been widely used in 3D self-supervised learning, achieving state-of-the-art performance. PointContrast \cite{xie2020pointcontrast} performs point-level invariant mapping on two different directional views of the same point cloud. It is the first unified framework to study the contrastive paradigm for 3D representation learning. STRL \cite{huang2021spatio} is an extension of the method \cite{grill2020bootstrap} on 3D point clouds to learn rich representational information by updating between online and target networks. PointSmile \cite{li2024pointsmile} guides network learning by maximizing the mutual information between two point clouds.  In addition, some researchers have focused on incorporating cross-model \cite{han2025cross,han2024trusted,hu2024hyperbolic,cheng2024multi,wu2023self,li2023cosine,zhang2023pointmcd,afham2022crosspoint} methods into discriminative methods. However, these methods often overlook the influence of local structures on whole shape responses and typically rely on large numbers of positive-negative sample pairs and complex strategies, resulting in a learning process that lacks physical interpretability.

\subsection{Physics-Informed Learning for Elasticity}
Recent physics-informed learning approaches \cite{kadambi2023incorporating} for elasticity often learn high-fidelity surrogate solvers for FEM-discretized fields by embedding physical laws into neural networks through loss terms or constraints. For example, DFEM \cite{xiong2025deep} couples PINNs with Finite Element Method (FEM) \cite{dhatt2012finite} discretization and leverages stiffness matrix constraints to improve the stability and accuracy of 3D elasticity solutions. Neural Modes \cite{wang2024neural}, on the other hand, learns an interpretable low-dimensional modal subspace by minimizing the system’s mechanical energy under equilibrium constraints. In the modeling of discrete structural systems, GNN-based methods (e.g., StructGNN-E \cite{song2023elastic} and Gulakala et al. \cite{gulakala2023graph}) further organize FEM-discretized nodes, edges, and boundary conditions into graphs, and use message passing to predict internal forces or von Mises stress fields. In contrast, we use static linear elasticity as a lightweight yet reliable physical prior to regularize point cloud self-supervised learning. Specifically, we combine data-driven learning with physics-based knowledge to capture how local deformations propagate to global responses under force equilibrium, thereby explicitly modeling the structural relationship between local parts and the whole object. This yields physically consistent and interpretable representations, mitigates the structural modeling limitations of purely data-driven methods, and improves robustness and generalization for downstream tasks.

\begin{figure*}[t]
	\centering
	\includegraphics[width=1\textwidth]{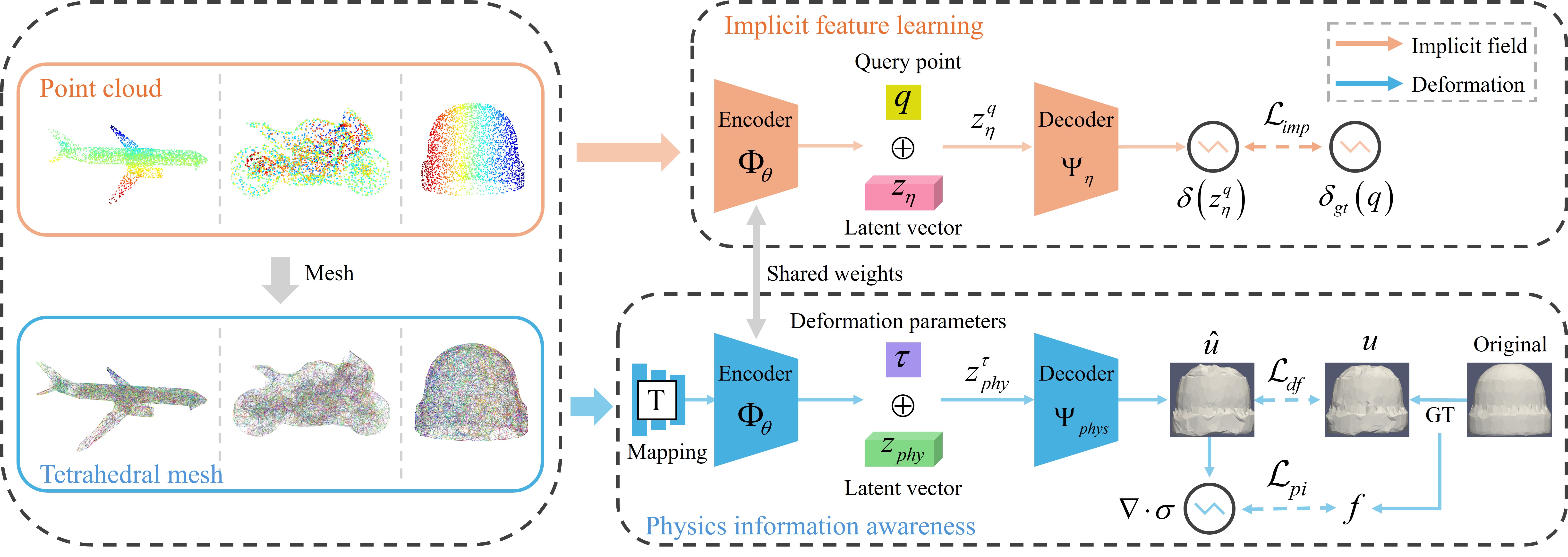}
	\caption{The overall architecture of the proposed method is illustrated. It consists of two modules: the implicit feature learning (IFL) module, which represents the point cloud as a continuous surface, helping the network better learn the whole shape features of the object; and the physics information awareness (PIA) module, which learns the relationship between local features and whole geometry through the elastic deformation of the object, enhancing the model’s ability to learn local details.}
	\label{fig2}
	\vspace{-1em}
\end{figure*}

\section{Method}
The framework of our method (Figure~\ref{fig2}) consists of two main modules. The first module is the implicit feature learning module, where the point cloud is input into an autoencoder to learn its implicit features. The implicit field represents the point cloud as a continuous surface, and by leveraging the distance function from a query point to the nearest surface, it can uniformly handle objects of arbitrary shapes, helping the network better learn the whole shape features of the object. The second module is the physical information awareness module, which captures local deformation behaviors under external forces by constructing tetrahedral meshes and introducing deformation mechanisms from elasticity theory. This module not only learns the physical characteristics of local responses, but also embeds local deformation features into a unified geometric representation space via a shared encoder, thereby enabling mutual enhancement between whole shape representation and local physical perception. The proposed dual-task self-supervised framework combines the interpretability of physics-based modeling with the strong representational capacity of data-driven methods, enabling more effective learning of the geometric representations of real-world objects.

\subsection{Preliminaries}
Given a point cloud ${P_i} = \left\{ {p_i^j \in {\mathbb{R}^3}|j = 1,2,...,n} \right\}$ representing a shape, we denote the implicit field query set as $Q = \left\{ {{q_k}} \right\}_{k = 1}^K$, where ${q_k} \in {\mathbb{R}}{^3}$. The backbone encoder ${{\Phi }_{\theta }}$ extracts a global latent code $z$ (dimension $m$), which is consumed by two decoders: an implicit decoder ${{\Psi }_{\eta}}$ for occupancy prediction and a deformation decoder ${{\Psi }_{phys}}$ for physics-driven shape response. For the physics branch, we represent the volumetric domain by a tetrahedral mesh ${\Omega _h}$ and use a mesh processor $\mathrm{T}$ to produce tetrahedron features compatible with ${{\Phi }_{\theta }}$, yielding a latent code ${z_{phys}}$. We then introduce $\tau$ as a physics-driven knowledge signal that defines the deformation configuration and thereby guides the network to learn physically meaningful deformation behavior. Subsequently, we concatenate $\tau$ with  ${z_{phys}}$ to form a conditioned representation (i.e., ${z_{phys}} \oplus d$), which is fed into ${{\Psi }_{phys}}$ to predict the displacement field and obtain the physics-driven shape response. In our implementation, $\tau$ encodes the key ingredients of linear elasticity, including material properties (e.g., Young’s modulus $E$ and Poisson’s ratio $\nu $, equivalently the Lamé parameters $\lambda$ and $\mu$), external loading specifications $f$ (e.g., body force and its direction), and boundary conditions (e.g., fixed region $\Gamma_D$). Details on the physical parameterization and setup are provided in the supplementary material.

\subsection{Implicit Feature Learning}
For a point cloud dataset $\mathcal{D}=\left\{ {{P}_{i}} \right\}_{i=1}^{\left| \mathcal{D} \right|}$, we use encoder ${{\Phi }_{\theta }}$ and decoder ${{\Psi }_{\eta}}$ to learn the implicit features of the point cloud ${P_i}$. The encoder ${{\Phi }_{\theta }}$ takes the ${P_i}$ as input and maps it to an $m$-dimensional latent vecoter ${z_\eta }$, represented as ${{\Phi }_{\theta }}:{{\mathbb{R}}^{n\times 3}}\to {{\mathbb{R}}^{m}},m\ll n$, where we use the network backbones of PointNet \cite{qi2017pointnet} or DGCNN \cite{wang2019dynamic} as ${{\Phi }_{\theta }}$ to learn the point cloud features, respectively. Subsequently, we concatenate latent vector ${z_\eta }$ with an arbitrary query point $q$ in 3D space to obtain feature vector $z_\eta ^q = {z_\eta } \oplus q$. Query points are randomly sampled within the normalized object bounding box, with an additional subset sampled near the surface to better capture geometric details. Then, the decoder ${{\Psi }_{\eta}}$ takes $z_\eta ^q$ as input and output an implicit function $\mathbb{S}={{\Psi }_{\eta}}\left( q,{{\Phi }_{\theta }}\left( {{P}_{i}} \right) \right)=\left\{ {\delta_i} |{\delta_i} \left( {z_\eta ^q} \right)\in \mathbb{R},\forall q\in {{\mathbb{R}}^{3}} \right\}$. ${\delta _i}\left( {z_\eta ^q} \right)$ represents the shortest distance from the query point $q$ to the point cloud ${P_i}$. The decoder ${{\Psi }_{\eta}}$ consists of 3-layer of MLP and outputs a 1-dimensional feature vector.

Our training objective is to minimize the error between the distance from the query point $q$ to the implicit surface of the point cloud ${P}_{i}$, as predicted by the implicit function $\delta_i({z_\eta ^q})$, and the ground truth distance ${\delta }_{gt}(q)$:

\begin{equation}
	{\theta ^*},{\eta ^*} = \mathop {\arg \min }\limits_{\theta ,\eta } \left( {\delta_i (z_\eta ^q)),{\delta _{gt}}\left( q \right)} \right),
\end{equation}

\noindent
where ${\delta }_{gt}(q)$ represents the nearest neighbor distance from the query point $q$ to the points on the point cloud ${{P}_{i}}$, expressed using the unsigned distance function value \cite{chibane2020neural}. Therefore, the implicit loss ${\cal L}_{imp}$ is defined as:

\begin{equation}
	{{\cal L}_{imp}} = \frac{1}{{|\mathcal{D}|K}}\sum\limits_{i = 1}^{|\mathcal{D}|} {\sum\limits_{q \in Q}^K {\left| {\left| {{\delta _i}(z_\eta ^q)} \right| - {\delta _{gt}(q)}} \right|} } ,
\end{equation}

\noindent
where $K$ is the number of the query point $q$ randomly sampled within the unit space, and $Q$ is the set of query points.

\subsection{Physical Information Awareness}
\begin{figure}[t]
	\centering
	\includegraphics[width=\linewidth]{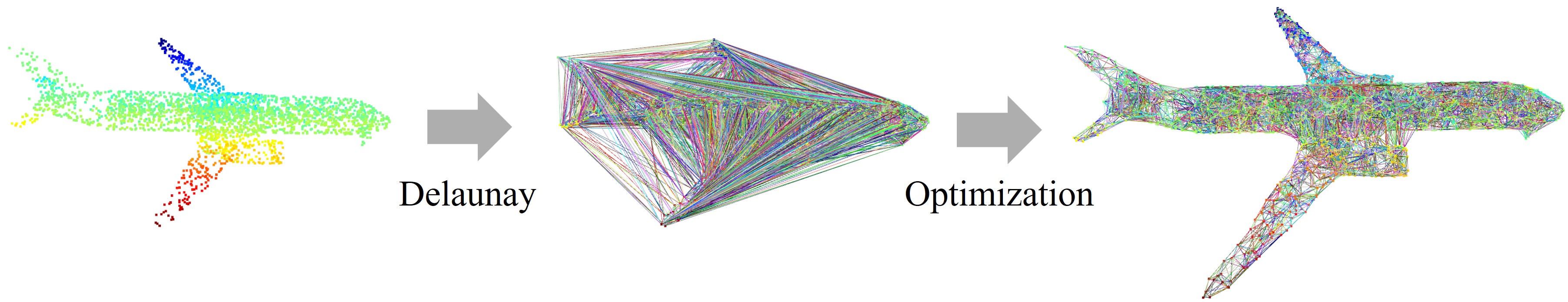}
	\caption{The point cloud is first converted into a coarse mesh form through Delaunay triangulation. Then, we removed the larger tetrahedral meshes, further refining them into the final tetrahedral representation.}
	\label{fig3}
	\vspace{-1em}
\end{figure}

For a point cloud ${{P }_{i }}$, we first use 3D Delaunay \cite{maur2002delaunay} to convert the ${{P }_{i }}$ into an object composed of multiple tetrahedra. It is an extension of Delaunay triangulation (which generates triangular meshes in 2D) to 3D space, producing a mesh composed of tetrahedral elements. Compared to hexahedral meshes, tetrahedral meshes better fit free-form objects and support complex topological structures, providing the volumetric elements required for physical modeling, such as deformation gradients and stress tensor computation. Meanwhile, directly using point coordinates lacks structural information, making it difficult to establish the strain–stress relationships necessary for constructing the subsequent physics-informed loss ${\mathcal{L}}_{pi}$. The advantages of using tetrahedral structures over point-based representations will be further validated in the ablation studies. The meshing process is shown in Figure~\ref{fig3}. 

Subsequently, the tetrahedral mesh is processed by the mesh processor $\mathrm{T}$, which encodes the local geometric and topological structure of each tetrahedron into a unified format suitable for input to the encoder ${{\Phi }_{\theta }}$. The mesh processor $\mathrm{T}$ consists of 3-layer of convolutional networks, bridging the mesh data and encoder ${{\Phi }_{\theta }}$. The encoder ${{\Phi }_{\theta }}$ maps the processed data to the $m$-dimensional latent space to obtain the latent vector ${z_{phys}}$. ${z_{phys}}$ is concatenated with the deformation parameters $\tau$ as the input $z_{phy}^\tau$ to decoder ${{\Psi }_{phys}}$, enabling the model to learn local-to-whole deformation patterns under varying physical conditions during training. The parameters in $\tau$ have clear physical meanings and characterize typical elastic responses under small deformations. Considering variations in point-cloud density and deformation scale, we further apply an adaptive scaling to $\tau$ based on the size of the generated tetrahedral mesh and the initial deformation magnitude, reducing manual tuning and improving robustness and generalization.The decoder ${{\Psi }_{phys}}$ is implemented as a five-layer MLP and outputs an $n\times 3$ displacement field, where $n$ denotes the number of input points.

Based on the configuration of deformation parameters $\tau$, we obtain the ground-truth displacement field $u$ using the FEM. Finally, based on the displacement field $\hat{u}$ predicted by the network and the ground-truth displacement field $u$ obtained through computation, we design two loss functions: the data-fidelity loss and the physics-informed loss. The data-fidelity loss guides the network to predict the displacement of each point in the deformed point cloud, explicitly modeling the relationship between local deformations and the whole geometric shape, thereby revealing the influence of local features on the whole structure. The physics-informed loss derives the external force distribution from the predicted local deformation data and compares it with the true external force distribution, ensuring that the network's output satisfies the physical law of force equilibrium. Below, we provide a detailed description of these two loss functions. 

\textbf{Data-fidelity Loss.} Its objective is to guide the network in accurately predicting the displacement field $\hat{u}$ deformed tetrahedral mesh object and comparing it with the ground truth displacement field  ${u}$, enabling it to precisely predict local geometric changes. For a tetrahedral mesh with undeformed nodal positions $x$, the deformed configuration is determined by node-wise displacements, i.e., $x' = x + u$. Since local geometric variations (e.g., stretching, compression, and bending patterns) arise from these nodal motions, estimating u provides an effective way to preserve geometric consistency of deformation at the element level. From the perspective of mechanics, forces are determined by deformation, and deformation is parameterized by the displacement field $u$. In the general dynamic setting, the deformation trajectory $x\left( t \right)$ can be obtained by extremizing the action functional:

\begin{equation}
	S\left( x \right)=\int_{0}^{T}{\left( \frac{1}{2}{{{\dot{x}}}^{T}}M\dot{x}-E\left( x \right)+{{x}^{T}}\left( {{f}_{e}}+{{f}_{d}} \right) \right)}dt,
\end{equation}

\noindent 
where $S(x)$ denotes the action functional of the system over the time interval $\left[ {0,T} \right]$. Here, $\frac{1}{2}{{{\dot{x}}}^{T}}M\dot{x}$ denotes the kinetic energy, with $M$ being the finite element mass matrix, $E(x)$ represents the potential energy (elastic deformation potential energy), and ${{x}^{T}}\left( {{f}_{e}}+{{f}_{d}} \right)$ corresponds to the external force action. ${{f}_{e}}$ are external forces and ${{f}_{d}}$ are dissipative frictional forces.

The above presents the general variational formulation for dynamic problems. To analyze deformation under static linear elasticity over the spatial domain ${\Omega }$, we typically solve for the displacement field $u$ by minimizing the total potential energy:

\begin{equation}
	S\left( u \right)=\int_{\Omega }{\left( \frac{1}{2}\sigma \colon \varepsilon \left( u \right)-f\cdot u \right)}d\Omega ,
\end{equation}

\noindent 
where ${\Omega }$ represents the volume region of the object. $S(u)$ is the total potential energy as a functional of the displacement field. $\frac{1}{2}\sigma \colon \varepsilon(u)$ represents the energy stored in the tetrahedral mesh due to elastic deformation. After the mesh is subjected to force, the resulting energy is calculated by the stress $\sigma$ and strain $\varepsilon$. $-f\cdot u$ represents the potential energy associated with external body forces under the displacement field, which is the work done by the external force under the displacement field $u$. Finally, we design the following data-fidelity loss:

\begin{equation}
	{{\rm{{\cal L}}}_{df}} = \frac{1}{{\left| {\rm{{\cal D}}} \right|M}}\sum\limits_{i = 1}^{\left| {\rm{{\cal D}}} \right|} {\sum\limits_{l = 1}^M {{{\left\| {{u_{il}} - {{\hat u}_{il}}} \right\|}_2}} } ,
\end{equation}

\noindent
where ${u_{il}}$ and ${{\hat u}_{il}}$ represent the ground truth displacement field of the tetrahedral mesh and the displacement field predicted by the network, respectively. $M$ denotes the number of tetrahedral meshes contained in the point cloud ${P_i}$. $\left\| \cdot \right\| _2$  represents the ${l_2}$-norm.

To ensure that the predicted deformation adheres to physical laws rather than merely fitting geometric targets, we introduce a physics-informed constraint governed by the fundamental force equilibrium equation \cite{dineva2019fundamental}. In static elasticity, for any volume element within the mesh to remain in a physically stable state, the divergence of its internal stress tensor $\sigma$ must balance the external body forces $f$. This governing equation is formally defined as: 

\begin{equation}
	\nabla \cdot \sigma + f = 0 \quad.
\end{equation}
 
Given the network predicted displacement field $\hat u$, we compute the stress field $\sigma$ and minimize the equilibrium residual $\left\| {\nabla  \cdot \sigma  + f } \right\|$ over all tetrahedra. This force-balance constraint couples neighboring elements and encourages globally consistent displacement predictions.

Specifically, we reconstruct the stress field $\sigma$ from the predicted vertex displacements using standard kinematics and Hooke’s law \cite{atanackovic2000hooke}. First, each tetrahedron consists of four vertices with initial coordinates ${{x}_{1}},{{x}_{2}},{{x}_{3}},{{x}_{4}}$ and deformed coordinates ${{{x}'}_{1}},{{{x}'}_{2}},{{{x}'}_{3}},{{{x}'}_{4}}$ (obtained from the network predictions). We construct the initial shape matrix ${X} = \left[ {{x_2} - {x_1}\quad{\rm{ }}{x_3} - {x_1}\quad{\rm{ }}{x_4} - {x_1}} \right]$ and the deformed shape matrix ${X'} = \left[ {{{x'}_2} - {{x'}_1}\quad{\rm{ }}{{x'}_3} - {{x'}_1}\quad{\rm{ }}{{x'}_4} - {{x'}_1}} \right]$. Then, we construct the deformation gradient $F$ based on the shape matrices $X$ and $X'$:

\begin{equation}
	F = X' \cdot {X^{ - 1}}
\end{equation}

After obtaining the deformation gradient, the strain tensor $\varepsilon$ of the tetrahedral mesh can be calculated using the following formula. $\varepsilon$ reflects the deformation information of each tetrahedron element, including both the stretching and shear components, as shown in the following equation:

\begin{equation}
	\varepsilon =\frac{1}{2}\left( F+{{F}^{T}} \right)-I ,
\end{equation}

\noindent
where ${F}^{T}$ denotes the transpose of the deformation gradien and $I$ is the identity matrix. Under linear elastic materials, the constitutive relation (Hooke’s law) yields the 3D stress tensor as:

\begin{equation}
	\sigma =\lambda Tr\left( \varepsilon  \right)I+2\mu \varepsilon ,
\end{equation}

\noindent
where $\lambda$ and $\mu$ is the Lamé constant. $Tr\left( \varepsilon  \right)$ is the trace of the strain tensor, representing the volumetric deformation. Finally, we define the physics-informed loss by averaging the ${l_2}$ norm of the equilibrium residual over all tetrahedra:

\begin{equation}
	{{\rm{{\cal L}}}_{pi}} = \frac{1}{{\left| D \right|M}}\sum\limits_{i = 1}^{\left| D \right|} {\sum\limits_{l = 1}^M {{{\left\| {\nabla  \cdot {\sigma _{il}} + {f_{il}}} \right\|}_2}} } ,
\end{equation}

\noindent
where ${\nabla  \cdot {\sigma _{il}}}$ represents the stress tensor divergence, which describes the force equilibrium within the mesh. $f_{il}$ denotes the ground-truth external force at the $l$-th tetrahedron of the $i$-th sample. This loss term evaluates whether each tetrahedral element satisfies physical equilibrium, thereby serving as a physics-based constraint to guide the network toward generating reasonable predictions that comply with mechanical laws.

\subsection{Overall Object}
Finally, we integrate the loss functions ${{\mathcal{L}}_{imp}}$, ${{\mathcal{L}}_{df}}$, and ${{\mathcal{L}}_{pi}}$ during the training phase, where ${{\mathcal{L}}_{imp}}$ represents the implicit loss, ${{\mathcal{L}}_{df}}$ represents the data-fidelity loss, and ${{\mathcal{L}}_{pi}}$ represents the physics-informed loss. $a$ and $b$ are coefficient.

\begin{equation}
	\mathcal{L}_{all}={{\mathcal{L}}_{imp}}+a{{\mathcal{L}}_{df}}+b{{\mathcal{L}}_{pi}}
\end{equation}

\section{Experiment}
\subsection{Pre-training}
\textbf{Baselines.} We compare our method with the current mainstream self-supervised learning methods: Jigsaw3D \cite{sauder2019self} and Rotation \cite{poursaeed2020self} are based on pretext tasks. CPG\cite{zhou2025cpg}, CCPoint \cite{11153700}, PointSmile \cite{li2024pointsmile}, PoCCA \cite{wu2024cross}, CLR-GAM \cite{malla2023clrgamcontrastivepointcloud}, ToThePoint \cite{li2023tothepoint}, and STRL \cite{huang2021spatio} are based on contrast learning methods. CrossCon-Jig \cite{han2025cross}, TCMSS \cite{han2024trusted}, HyperIPC\cite{hu2024hyperbolic}, MCIB\cite{cheng2024multi}, CrossNet \cite{wu2023self}, Cosine Mixup \cite{li2023cosine}, PointMCD \cite{zhang2023pointmcd}, and CrossPoint \cite{afham2022crosspoint} are based on cross-modal methods. MAE3D \cite{10250984}, GDANet\cite{li2025unified}, GSPCon \cite{han2023graph}, Tsai et al. \cite{tsai2022self}, and OcCo \cite{wang2021unsupervised} are based on the generative models. Additionally, for clarity in the tables below, we use the abbreviations Sup, PT, CL, CM, and GM to represent supervised learning, pretext tasks, contrastive learning, cross-modal methods, and generative models, respectively.

\textbf{Datasets.} We use the ShapeNet \cite{chang2015shapenet} dataset as the pre-training dataset, which consists of 57,448 synthetic models from 55 categories. For downstream tasks, we evaluate our method on four different benchmark datasets for object classification and object segmentation. Object Classification: the synthetic dataset ModelNet40 \cite{wu20153d} and the real-world dataset ScanObjectNN \cite{uy2019revisiting}. Object Segmentation: the synthetic dataset ShapeNetPart \cite{yi2016scalable} and the real-world dataset S3DIS \cite{armeni20163d}.

\begin{table}[!t]
	\caption{Linear classification accuracy (\%) on ModelNet40 and ScanObjectNN. PN and DG denote PointNet and DGCNN backbones respectively.}
	\label{table1}
	\centering
	\setlength{\tabcolsep}{4.5pt} 
	\renewcommand{\arraystretch}{1.1}
	\begin{tabular}{l|c|cc|cc}
		\toprule
		\multirow{2}{*}{\textbf{Method}} & \multirow{2}{*}{\textbf{Category}} 
		& \multicolumn{2}{c|}{\textbf{ModelNet40}} & \multicolumn{2}{c}{\textbf{ScanObjectNN}} \\
		\cline{3-6}
		& & \textbf{PN} & \textbf{DG} & \textbf{PN} & \textbf{DG} \\ 
		\hline
		PointNet \cite{qi2017pointnet} &\multirow{2}{*}{Sup} & 89.2 & $-$ & 73.3 & $-$ \\ 
		DGCNN \cite{wang2019dynamic} &  & $-$ & 92.9 & $-$ & 82.8 \\ 
		\hline
		Jigsaw3D \cite{sauder2019self} &\multirow{2}{*}{PT} & 87.3 & 90.6 & 55.2 & 59.5 \\ 
		Rotation \cite{poursaeed2020self} &  & 88.6 & 90.8 & $-$ & $-$ \\ 
		\hline
		STRL \cite{huang2021spatio} & \multirow{6}{*}{CL} & 88.3 & 90.9 & 74.2 & 77.9 \\
		ToThePoint \cite{li2023tothepoint} &  & 85.6 & 89.2 & 74.7 & 81.9 \\
		CLR-GAM \cite{malla2023clrgamcontrastivepointcloud} &  & 88.9 & 91.3 & 75.7 & 82.1 \\
		PoCCA \cite{wu2024cross} &  & 89.4 & 91.4 & 75.6 & 82.2 \\
		PointSmile \cite{li2024pointsmile} &  & 90.0 & 91.8 & 75.8 & 82.8 \\
		CCPoint \cite{11153700} &  & \underline{90.2}  & \textbf{92.4} & $-$ & \underline{86.2} \\
		\hline
		CrossPoint \cite{afham2022crosspoint} & \multirow{7}{*}{CM} & 89.1 & 91.2 & 75.6 & 81.7 \\
		Cosine Mixup \cite{li2023cosine} &  & 88.7 & 91.7 & 72.5 & 84.7 \\
		CrossNet \cite{wu2023self} &  & 89.5 & 91.5 & \underline{76.8} & 83.9 \\
		HyperIPC \cite{hu2024hyperbolic} &  & $-$ & 91.8 & $-$ & 84.5 \\
		TCMSS \cite{han2024trusted} &  & 87.7 & 91.4 & 75.1 & 84.0 \\
		MCIB \cite{cheng2024multi} &  & $-$ & 91.6 & 76.6 & 85.4 \\
		CrossCon-Jig \cite{han2025cross} &  & 89.3 & 92.0 & 75.9 & 83.6 \\
		\hline
		OcCo \cite{wang2021unsupervised} & \multirow{5}{*}{GM} & 88.7 & 89.2 & 69.5 & 78.3 \\ 
		Tsai et al. \cite{tsai2022self} &  & 90.1  & 92.0 & 70.6 & 74.8 \\
		GSPCon \cite{han2023graph} &  & 89.8 & 91.5 & 75.8 & 83.9 \\
		MAE3D \cite{10250984} &  & $-$ & \textbf{92.4} & $-$ & $-$ \\
		\textbf{Ours} &  & \textbf{90.3} & \underline{92.2} & \textbf{79.5} & \textbf{86.4} \\
		\bottomrule
	\end{tabular}
	\vspace{-1em}
\end{table}

\textbf{Implementation Details.} To facilitate a fair comparison with existing self-supervised learning methods, we follow the experimental setup proposed in \cite{huang2021spatio} and use the backbone networks of PointNet \cite{qi2017pointnet} or DGCNN \cite{wang2019dynamic} as the encoder ${{\Phi }_{\theta }}$. We use Adam \cite{kingma2014adam} optimizer with weight decay $1\times {{10}^{-4}}$ and initial learning rate $1\times {{10}^{-3}}$. The learning rate scheduler is cosine annealing \cite{loshchilov2016sgdr}, and the model is trained end-to-end across 200 epochs. After pre-training, all downstream tasks are performed on the pre-trained encoder ${{\Phi }_{\theta }}$. Our models are trained on two Tesla V100-PCI-E-32G GPUs. For the classification task, we use a pre-trained backbone model to extract point cloud features on the ModelNet40 dataset and ScanObjectNN dataset, and these features are used to train a simple linear Support Vector Machine (SVM) classifier \cite{suthaharan2016support}. For segmentation tasks, We evaluate part segmentation using the mean intersectionover-union (mIoU) metric, which is computed by averaging the IoU of each part in the object and then averaging the values obtained from each object category. In dataset S3DIS, we performed 6-fold cross-validation on its 6 areas to obtain the final segmentation results. Additional experiments are provided in the supplementary material.

\subsection{Object Classification}
\textbf{Linear Classification.} We conduct linear classification experiments on the ModelNet40 and ScanObjectNN datasets to evaluate the representation capability of the pretrained feature extractor. As shown in Table~\ref{table1}, the proposed method achieves achieves excellent linear classification accuracy on the ModelNet40 dataset, with 90.3$\%$ using PointNet and 92.2$\%$ using DGCNN. In addition, on the ScanObjectNN dataset, our method achieves 79.5$\%$ and 86.4$\%$accuracy under the PointNet and DGCNN backbones, respectively, both outperforming other methods with improvements of 2.7$\%$ and 0.2$\%$. Notably, on the ScanObjectNN dataset, our method achieves improvements of 6.2$\%$ and 3.6$\%$ over the original supervised learning models PointNet and DGCNN, respectively, demonstrating strong generalization capability in real-world scenarios. This indicates that our physically driven structural modeling effectively enhances the expressiveness of point cloud features, leading to superior classification performance.

\begin{figure}[!t]
	\centering
	\includegraphics[width=\linewidth]{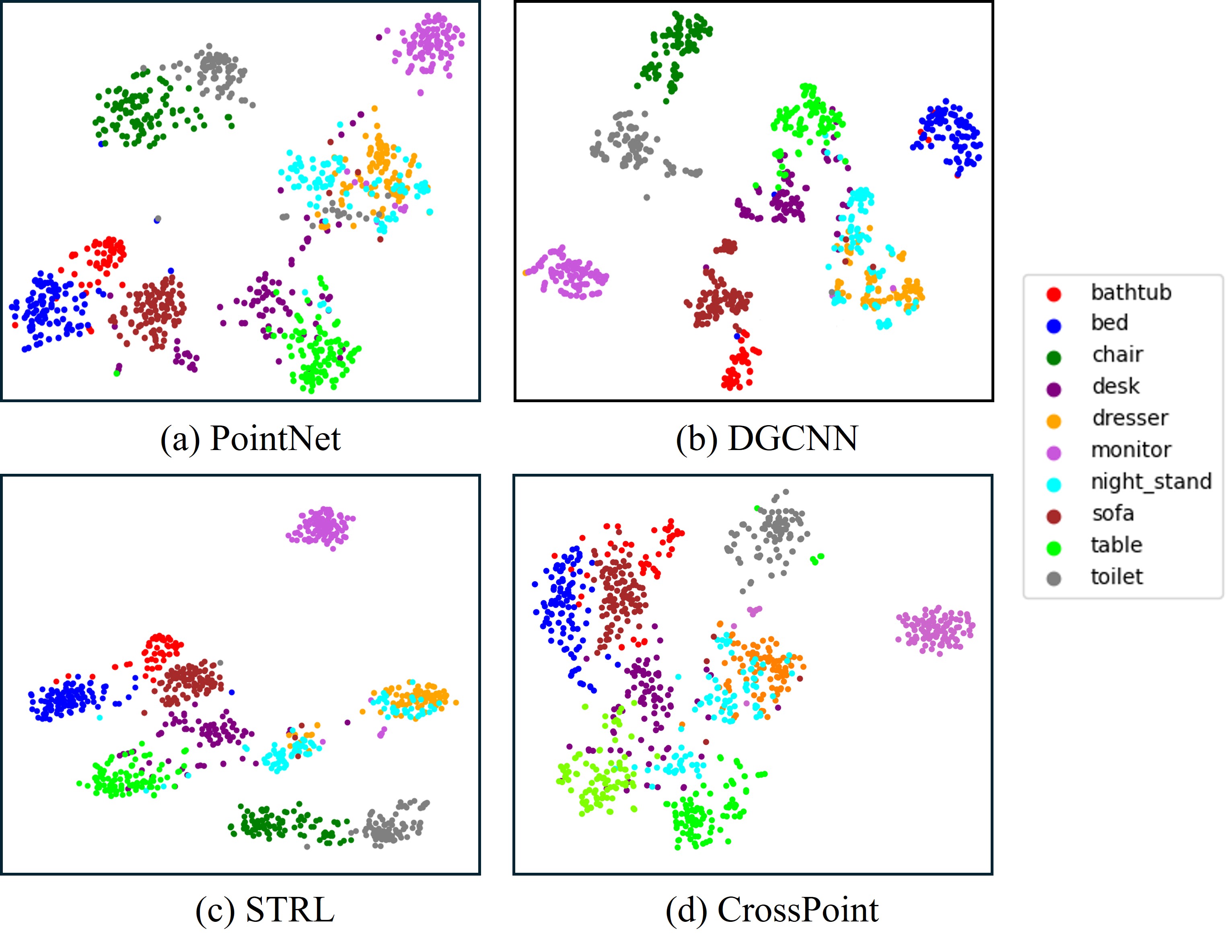}
	\caption{Feature-space visualization (t-SNE) of learned point-cloud embeddings on ShapeNet with representative baselines. (a) Ours with PointNet backbone, (b) Ours with DGCNN backbone, (c) STRL (DGCNN), and (d) CrossPoint (DGCNN). Colors indicate object categories.}
	\label{fig4}
	\vspace{-1em}
\end{figure}

We evaluate the trained pre-trained weights on the ModelNet10 dataset by extracting features for each sample and performing t-SNE \cite{van2008visualizing} clustering visualization, as shown in Figure~\ref{fig4}. (a) and (b) show the feature distributions obtained by our method with PointNet and DGCNN backbones, respectively, while (c) and (d) correspond to two representative baselines: the contrastive learning method STRL (with DGCNN backbone) and the cross-modal method CrossPoint (with DGCNN backbone). As can be seen, our method forms more compact intra-class clusters and clearer inter-class separation under both backbones. Most categories exhibit well-defined clustering boundaries and fewer outliers, indicating more stable representational capabilities (the consistent trend from (a) to (b) further suggests that our method is relatively robust to the choice of backbone). In contrast, the feature distributions of STRL and CrossPoint tend to show more intra-class dispersion and elongated cluster shapes, suggesting that the representations they learn are less robust in terms of structural consistency. Our approach is capable of learning more structured point cloud representations.

\begin{table}[!t]
	\caption{Fine-tuning accuracy (\%) of shape classification on the ModelNet40 dataset and the ScanObjectNN dataset. The self-supervised pre-trained model was used as the initial weights for the supervised learning method. PN and DG denote PointNet and DGCNN.}
	\label{table2}
	\begin{center}
		\begin{small}
			\setlength{\tabcolsep}{3.5pt}
			\renewcommand{\arraystretch}{1.1}
			\begin{tabular}{l|c|cc|cc}
				\toprule
				\multirow{2}{*}{\textbf{Method}} & \multirow{2}{*}{\textbf{Category}} 
				& \multicolumn{2}{c|}{\textbf{ModelNet40}} & \multicolumn{2}{c}{\textbf{ScanObjectNN}} \\
				\cline{3-6}
				& & \textbf{PN} & \textbf{DG} & \textbf{PN} & \textbf{DG} \\
				\hline
				PointNet \cite{qi2017pointnet}  &\multirow{2}{*}{Sup} & 89.2 & $-$ & 73.3 & $-$ \\
				DGCNN \cite{wang2019dynamic} &  & $-$ & 92.9 & $-$ & 82.8 \\
				\hline
				Jigsaw3D \cite{sauder2019self} & PT & 89.6 & 92.4 & 76.5 & 82.7 \\
				\hline
				STRL \cite{huang2021spatio} &\multirow{5}{*}{CL} & $-$ & 93.1 & $-$ & $-$ \\
				PoCCA \cite{wu2024cross} &  & 90.2 & 93.2 & \underline{80.3} & \underline{84.1} \\
				PointSmile \cite{li2024pointsmile} &  & \textbf{90.7} & 93.0 & $-$ & $-$ \\
				CPG \cite{zhou2025cpg} &  & $-$ & 93.6 & $-$ & $-$ \\
				CCPoint \cite{11153700} &  & $-$ & 93.4 & $-$ & $-$ \\
				\hline
				PointMCD \cite{zhang2023pointmcd}  &\multirow{2}{*}{CM} & 91.1 & \underline{93.7} & $-$ & $-$ \\
				CrossNet \cite{wu2023self} &  & $-$ & 93.4 & $-$ & $-$ \\
				\hline
				OcCo \cite{wang2021unsupervised} &\multirow{4}{*}{GM} & 90.1 & 93.1 & 80.0 & 83.9 \\
				GDANet \cite{li2025unified} &  & 90.1 & 93.3 & 69.0 & 79.4 \\
				MAE3D \cite{10250984} &  & \underline{90.6} & 93.4 & $-$ & $-$ \\
				\textbf{Ours} &  & \underline{90.6} & \textbf{93.9} & \textbf{80.9} & \textbf{88.0} \\
				\bottomrule
			\end{tabular}
		\end{small}
	\end{center}
	\vspace{-1em}
\end{table}

\textbf{Supervised Fine-tuning.} We fine-tune our model on ModelNet40 and ScanObjectNN, as shown in Table~\ref{table2}. Compared with existing methods, ours achieves overall superior performance, outperforming the original supervised learning baselines (PointNet and DGCNN). Notably, our method achieves comparable performance to the best existing method on ModelNet40, while demonstrating the most significant improvement on the real-world ScanObjectNN dataset when using the DGCNN architecture, reaching 88.0$\%$, which is 3.9$\%$ higher than the second-best method (PoCCA  \cite{wu2024cross}); under PointNet, it exceeds the second-best by 0.6$\%$. These results indicates that our method generalizes well even in challenging scenarios. These findings demonstrate that incorporating physical information not only enhances the model's understanding of real-world objects, but also effectively establishes structural dependencies between local and whole regions through the mechanism of force propagation. This significantly improves the model’s representational learning capacity and generalization performance across different point cloud datasets.

\begin{table}[!t]
	\caption{Performance comparison of segmentation under ShapeNetPart dataset and S3DIS dataset (DGCNN as backbone).}
	\label{table3}
	
	\begin{center}
		\begin{small}
			\setlength{\tabcolsep}{5pt} 
			\renewcommand{\arraystretch}{1.1}
			\begin{tabular}{l|l|ccr}
				\toprule
				\multirow{2}{*}{\textbf{Method}} & \multirow{2}{*}{\textbf{Category}} & \multirow{2}{*}{\textbf{ShapeNetPart}} & \multirow{2}{*}{\textbf{S3DIS}} \\ 
				& & & \\ \hline
				DGCNN \cite{wang2019dynamic} & \multicolumn{1}{c|}{\multirow{1}{*}{Sup}} & 85.1 & 54.9 \\ \hline 
				
				Jigsaw3D \cite{sauder2019self} & \multicolumn{1}{c|}{\multirow{1}{*}{PT}} & 84.3 & 55.6 \\ \hline
				
				STRL \cite{huang2021spatio} & \multicolumn{1}{c|}{\multirow{6}{*}{CL}} & 85.1 & $ - $ \\
				ToThePoint \cite{li2023tothepoint} & \multicolumn{1}{c|}{} & 85.5 & $ - $ \\
				PointSmile \cite{li2024pointsmile} & \multicolumn{1}{c|}{} & 85.4 & \underline{58.9} \\
				CLR-GAM \cite{malla2023clrgamcontrastivepointcloud} & \multicolumn{1}{c|}{} & 85.5 & $ - $ \\
				\multicolumn{1}{l|}{PoCCA  \cite{wu2024cross}} & \multicolumn{1}{c|}{} & 85.8 & $ - $ \\
				CCPoint \cite{11153700} & \multicolumn{1}{c|}{} & 85.6 & $-$ \\ 
				\hline
				
				CrossPoint \cite{afham2022crosspoint} &\multicolumn{1}{c|}{\multirow{6}{*}{CM}} & 85.3 & 58.4 \\
				\multicolumn{1}{l|}{ PointMCD \cite{zhang2023pointmcd}} & \multicolumn{1}{c|}{} & 85.1 & $ - $\\
				\multicolumn{1}{l|}{ TCMSS \cite{han2024trusted}} & \multicolumn{1}{c|}{} & 85.8 & 58.8\\
				CrossNet \cite{wu2023self} & \multicolumn{1}{c|}{} & 85.5 & $ - $ \\
				MCIB \cite{cheng2024multi} & \multicolumn{1}{c|}{} & \textbf{86.2} & $ - $ \\
				CrossCon-Jig \cite{han2025cross} & \multicolumn{1}{c|}{} & 85.6 & $ - $ \\
				
				\hline
				
				OcCo \cite{wang2021unsupervised} & \multicolumn{1}{c|}{\multirow{6}{*}{GM}} & 85.0 & 58.0 \\ 
				\multicolumn{1}{l|}{ GSPCon  \cite{han2023graph}} & \multicolumn{1}{c|}{} & 85.7 & $ - $ \\
				\multicolumn{1}{l|}{ Tsai et al.  \cite{tsai2022self}} & \multicolumn{1}{c|}{} & 85.1 & $ - $ \\
				\multicolumn{1}{l|}{ GSPCon  \cite{han2023graph}} & \multicolumn{1}{c|}{} & 85.7 & $ - $ \\
				\multicolumn{1}{l|}{ GDANet  \cite{li2025unified}} & \multicolumn{1}{c|}{} & $ - $ & \textbf{59.0} \\
				\textbf{Ours} & \multicolumn{1}{c|}{} & \underline{86.0} & \textbf{59.0}\\
				\bottomrule
			\end{tabular}
		\end{small}
	\end{center}
	\vspace{-1em}
\end{table}

\begin{figure*}[!t]
	\centering
	\includegraphics[width=\linewidth]{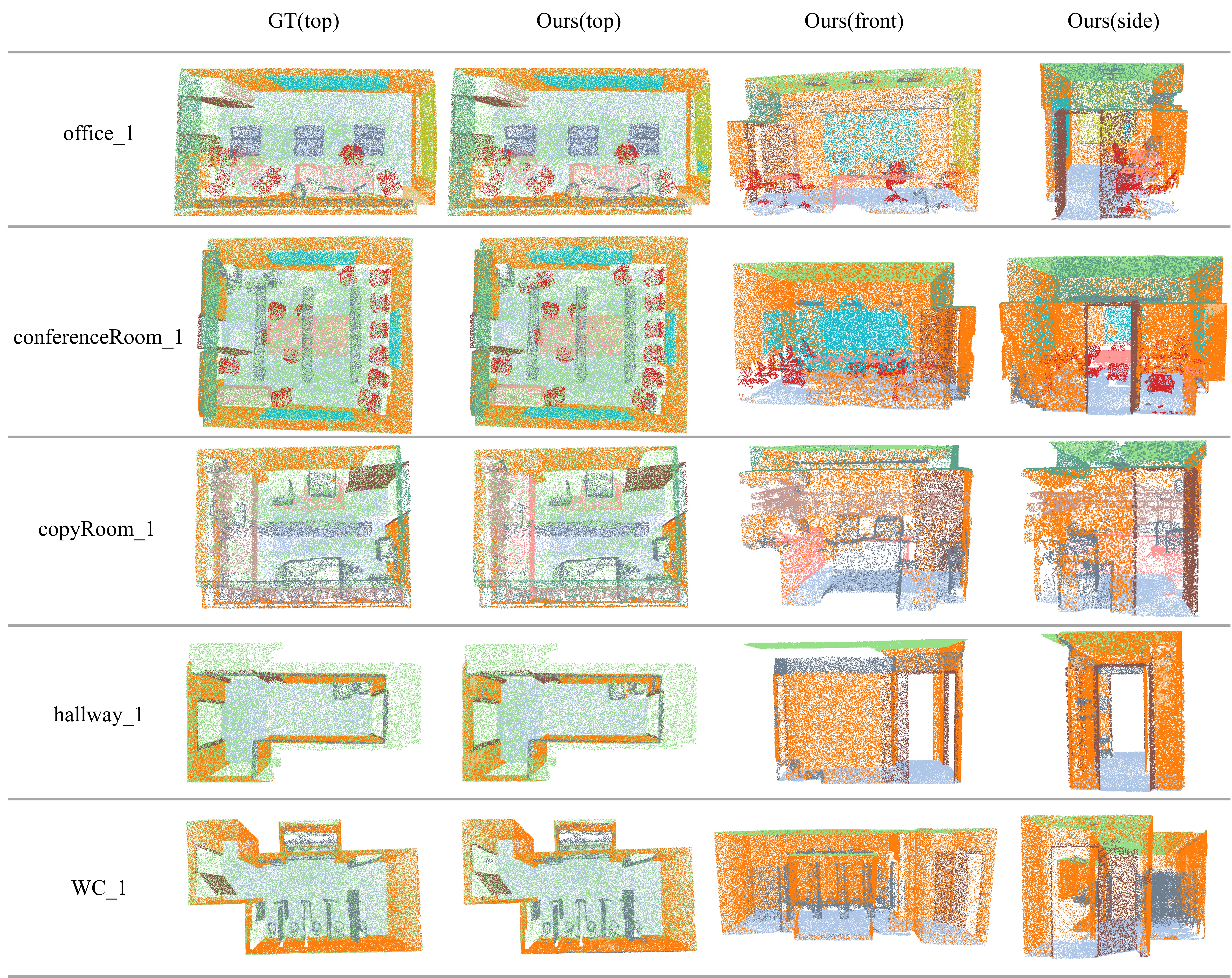}
	\caption{Semantic segmentation results on S3DIS (DGCNN as backbone). Different colors indicate different objects. From left to right are the top view of the ground truth, the predicted top view, the front view, and the side view.}
	\label{fig5}
	\vspace{-1em}
\end{figure*}

\begin{figure}[!t]
	\centering
	\includegraphics[width=\linewidth]{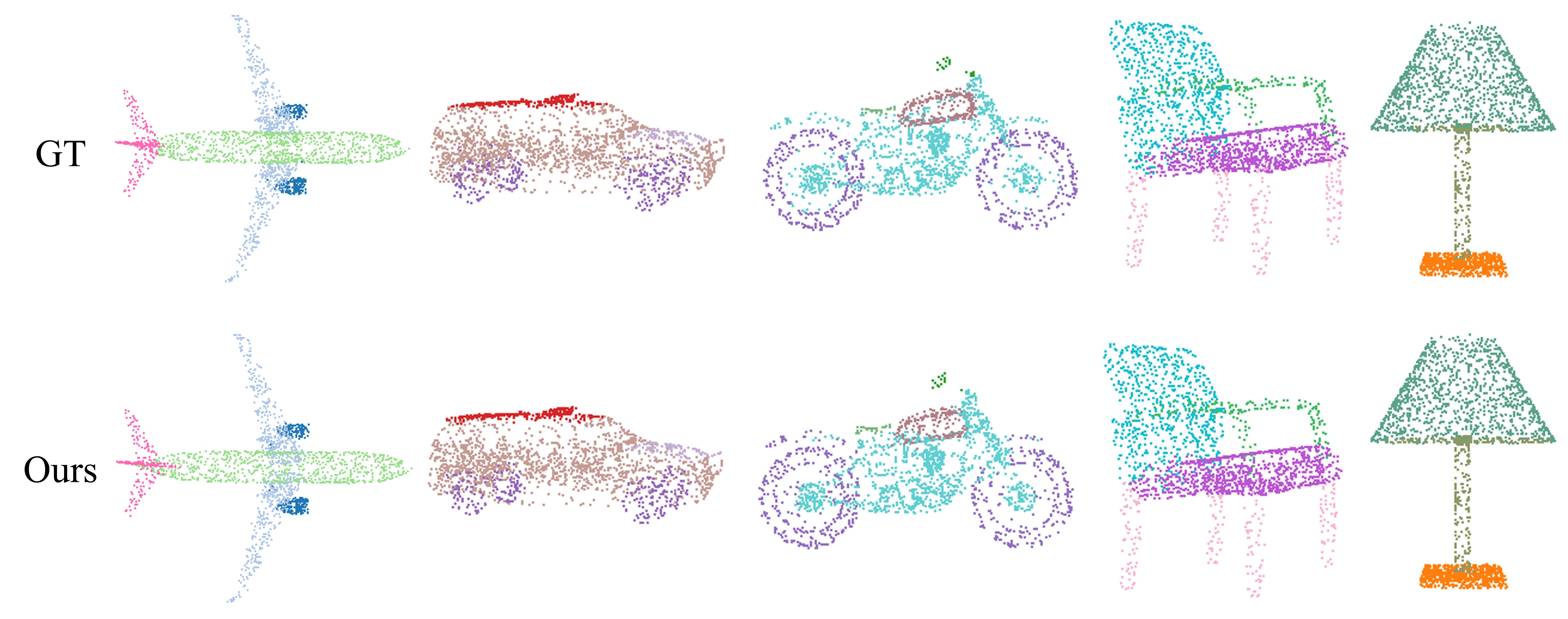}
	\caption{Visualisation of part segmentation results based on DGCNN backbone. Different colours indicate different parts. Top row: ground-truth; bottom row: The results of our projections.}
	\label{fig6}
	\vspace{-1em}
\end{figure}

\subsection{Object Segmentation}
To further validate the effectiveness of the physical mechanism in point cloud representation learning, we evaluate our method on object segmentation task. This task requires the model not only to understand the global geometric structure, but also to accurately capture the relative relationships between individual parts. We transfer the pre-trained DGCNN backbone model to the segmentation task and conduct experimental analysis using the ShapeNetPart and S3DIS datasets, fine-tuning the model in an end-to-end manner on both datasets. We report the mean IoU (Intersection-over-Union) metric in Table~\ref{table3}. Specifically, we perform 6-fold cross-validation on the 6 regions of the S3DIS dataset to obtain the final segmentation results. It can be seen that our method performs well on both datasets, demonstrating that the incorporation of physical information enhances the model's segmentation capability. Figure~\ref{fig5} and Figure~\ref{fig6} show the segmentation visualizations on the S3DIS and the ShapeNetPart dataset dataset, respectively. It can be observed that by introducing a force-driven local-to-global modeling mechanism, our method exhibits strong generalization ability across different fine-grained point cloud segmentation tasks, demonstrating a solid understanding of local details as well as the geometric relationships and structural layouts in complex indoor scenes.

\subsection{Ablation Study}

\textbf{Module Analysis.} we conducted a qualitative analysis of the performance impact of each part on the ModelNet40 and ScanObjectNN datasets. Following the same protocol as the classification experiments, we pre-train the backbone under three module settings: Physical Information Awareness (PIA), Implicit Feature Learning (IFL), and their combination (PIA+IFL). We then train a linear SVM on the extracted point-cloud features, as summarized in Table~\ref{table4}. The results demonstrate that incorporating physical information as a constraint into the model learning process enhances the detection performance, enabling the model to learn more discriminative point cloud representations.

\begin{figure*}[!t]
	\centering
	\includegraphics[width=1\linewidth]{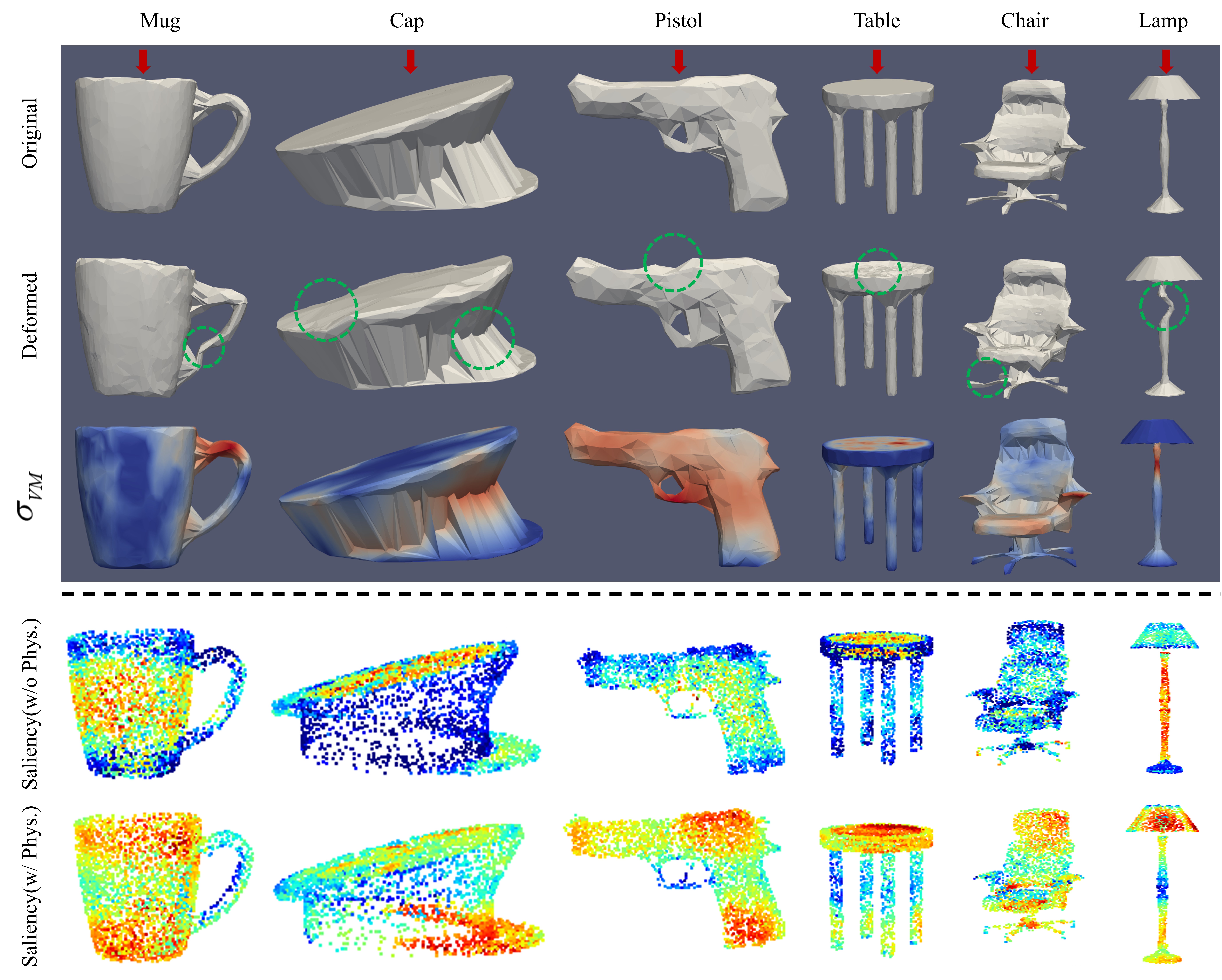}
	\caption{Visualization of deformation and saliency feature. From top to bottom: (1) original shapes; (2) predicted elastic deformations under the same downward loading; (3) visualization of the load response (von Mises equivalent stress ${\sigma _{VM}}$ computed from the displacement field is rendered to characterize the spatial distribution of the load response, with colors varying from cool to warm indicating increasing response magnitude); (4)(5) Point-wise saliency heatmaps generated by the models without and with physical constraints, respectively, used to compare the differences in attended regions under the two settings (colors from cool to warm indicate increasing saliency). The red arrows indicate the direction of the applied force, while the green dashed circles highlight regions exhibiting significant deformation.}
	\label{fig7}
	\vspace{-1em}
\end{figure*}

\begin{table}[!t]
	\caption{Classification results using linear SVM for pre-trained models (DGCNN backbone) on ModelNet40 and ScanObjectNN datasets.}
	\label{table4}
	\begin{center}
		\begin{small}
			\setlength{\tabcolsep}{9pt} 
			\renewcommand{\arraystretch}{1.3}
			\begin{tabular}{c|l|c|cr}
				\toprule
				\multirow{2}{*}{\textbf{Datasets}} & \multirow{2}{*}{PIA} & \multirow{2}{*}{IFL} & \multirow{2}{*}{PIA+IFL} \\ 
				& & & \\ \hline
				ModelNet40 & \multicolumn{1}{c|}{\multirow{1}{*}{85.1}} & 90.5 & 92.2 \\ \hline 
				
				ScanObjectNN & \multicolumn{1}{c|}{\multirow{1}{*}{78.3}} & 84.1 & 86.4 \\ 
				\bottomrule
			\end{tabular}
		\end{small}
	\end{center}
	\vspace{-1em}
\end{table}

\begin{table}[!t]
	\caption{Classification results using linear SVM for pre-trained models (DGCNN backbone) on ModelNet40 dataset.}
	\label{table5}
	\begin{center}
		\begin{small}
			\setlength{\tabcolsep}{5pt} 
			\renewcommand{\arraystretch}{1.3}
			\begin{tabular}{c|l|c|c|c|cr}
				\toprule
				\multirow{2}{*}{\textbf{Test}} & \multirow{2}{*}{point cloud} & \multirow{2}{*}{ tetrahedral} & \multirow{2}{*}{${{\cal L}_{df}}$} & \multirow{2}{*}{ ${{\cal L}_{pi}}$} & \multirow{2}{*}{ accuracy} \\ 
				& & & & \\ \hline
				1 & \multicolumn{1}{c|}{\multirow{1}{*}{}} & \checkmark & \checkmark & & 91.7\\ \hline 
				2 & \multicolumn{1}{c|}{\multirow{1}{*}{}} & \checkmark & & \checkmark & 91.3\\ \hline 
				3 & \multicolumn{1}{c|}{\multirow{1}{*}{\checkmark}} & & \checkmark & &90.9\\ \hline 
				4 & \multicolumn{1}{c|}{\multirow{1}{*}{}} & \checkmark & \checkmark& \checkmark & \textbf{92.2}\\ 
				\bottomrule
			\end{tabular}
		\end{small}
	\end{center}
	\vspace{-1em}
\end{table}

\textbf{Point Cloud VS Tetrahedral Structure.} We further analyze the importance of directly inputting the tetrahedral mesh into the physics information awareness module, as shown in Table~\ref{table5}. Since point clouds inherently lack explicit topological structure, the stress–strain relationships required by the physics-informed loss term $\mathcal{L}_{\text{pi}}$ cannot be established when using only the point cloud as input. This leads to the failure of the intended physical constraints. A comparison between Test 1, Test 2, and Test 4 shows that incorporating either $\mathcal{L}_{\text{df}}$ or $\mathcal{L}_{\text{pi}}$ leads to accuracy improvements, validating the positive role of both loss terms in modeling the relationship between local and whole physical responses. Furthermore, comparing Test 3 (point cloud only) with Test 1 (tetrahedral mesh) reveals that the former achieves an accuracy of 90.9$\%$, slightly lower than the 91.7$\%$ of the latter. This indicates that introducing structured tetrahedral elements more effectively captures local topological relationships, thereby enhancing the network’s ability to model deformation behavior. In summary, structured input is not only a necessary prerequisite for physical information modeling but also contributes directly and positively to the final recognition performance.

\textbf{Deformation visualization and saliency analysis.} To further provide an intuitive analysis of the model’s response patterns to deformation, we jointly visualize the predicted deformations and feature saliency. As shown in Figure~\ref{fig7}, in the first three rows, the most pronounced local deformations are concentrated in structurally critical regions (e.g., mug handles, hat brims, pistol barrels, lamp arm joints, and load-bearing surfaces of chairs). These regions are typically stress-sensitive and therefore more prone to stress concentration (highlighted by green dashed circles). Importantly, these local deformations do not occur in isolation but arise as a result of under the influence of global force propagation through the structure, reflecting the network's ability to model the physical coupling between local parts and the global shape. Comparing with the fourth row of Figure~\ref{fig7} (the variant without physical constraints), we observe that its saliency responses tend to be more scattered and are more easily affected by local geometric details or noise. In contrast, the saliency maps in the fifth row (with the physics-based mechanism enabled) are more strongly focused on the primary load-bearing regions and their neighboring load-transfer paths, and exhibit a much clearer spatial correspondence with the stress distribution. These results indicate that the physics-driven mechanism guides the network to shift its attribution from dispersed local geometric perturbations toward key load-bearing areas, thereby more faithfully capturing the propagation from local force responses to global deformations and improving the consistency and interpretability of the local-to-whole modeling.

Due to space limitations, additional ablation and analysis results are deferred to the supplementary material. Specifically, we include (i) qualitative module analysis; (ii) a hyperparameter study of the loss weights $a$ and $b$; (iii) implementation details and the physical configuration of the FEM-based module (parameterization and pseudocode); (iv) tetrahedral mesh generation and quality analysis, covering volume-based filtering as well as geometric fidelity and element quality for the Delaunay mesh; and (v) a training effectiveness analysis, reporting preprocessing overhead and training-time performance comparisons.

\section{Conclusion}
This paper proposes a physics-driven self-supervised learning method for point cloud representation, which effectively models structural responses under force propagation by introducing a local-whole deformation mechanism derived from elasticity theory. The framework consists of two training modules: one models the whole geometric structure using an implicit field, enabling expressive representation for arbitrary shapes; the other captures elastic responses induced by external forces based on tetrahedral mesh construction and constrains deformation behavior through a physics loss. Both branches share encoder weights, establishing a closed-loop mapping between local force-induced deformations and global shape changes, thereby deeply integrating local details with global structural features. Experimental results demonstrate that the proposed method achieves superior performance across multiple downstream tasks, including classification and segmentation. These results validate the effectiveness of incorporating physics-based modeling for enhancing the robustness and generalization capability of point cloud representations.

\bibliographystyle{IEEEtran}
\bibliography{IEEEabrv,reference}

\end{document}